# Adversarial Networks for Prostate Cancer Detection

**Simon Kohl**
Medical Image Computing
German Cancer Research Center (DKFZ)
Heidelberg, Germany
`simon.kohl@dkfz.de`

**David Bonekamp**
Department of Radiology, DKFZ

**Heinz-Peter Schlemmer**
Department of Radiology, DKFZ

**Kaneschka Yaqubi**
Department of Radiology, DKFZ

**Markus Hohenfellner**
Department of Urology
Heidelberg University Hospital, Germany

**Boris Hadaschik**
Department of Urology
Essen University Hospital, Germany

**Jan-Philipp Radtke**
Department of Radiology, DKFZ,
Department of Urology
Heidelberg University Hospital, Germany

**Klaus Maier-Hein**
Medical Image Computing, DKFZ

## Abstract

The large number of trainable parameters of deep neural networks renders them inherently data hungry. This characteristic heavily challenges the medical imaging community and to make things even worse, many imaging modalities are ambiguous in nature leading to rater-dependant annotations that current loss formulations fail to capture. We propose employing adversarial training for segmentation networks in order to alleviate aforementioned problems. We learn to segment aggressive prostate cancer utilizing challenging MRI images of 152 patients and show that the proposed scheme is superior over the de facto standard in terms of the detection sensitivity and the dice-score for aggressive prostate cancer. The achieved relative gains are shown to be particularly pronounced in the small dataset limit.

## 1 Introduction

The standard approach to training fully convolutional networks (FCNs) relies on a per-pixel formulation of the loss, treating individual pixels in the label maps as conditionally independent from all others. While this approach works for scenes with well-bounded objects, heterogeneous and amorphous structures are abundant and often of prime interest in medical images. To accommodate for the ensuing ambiguities in the manually generated groundtruth labels, we propose to follow a recent line of work that employs adversarial networks for segmentation [1, 2] with the rationale that the adversary allows for multiple equally plausible segmentations without arbitrarily favoring one over the other. We test this hypothesis in the context of the segmentation and thus detection of aggressive prostate cancer (PC) from MRI. In contrast to the large body of studies for the delineation of PC from MRI that mostly operate voxel-wise on pre-segmented regions of interest, e.g. the prostate as a whole ([3, 4, 5, 6]), we propose a joint FCN-based segmentation of the prostate's regions along with the targeted cancer nodules that is learned end-to-end using *purely* adversarial training.

## 2 Methods

### 2.1 Adversarial Training for Semantic Segmentation

The general idea behind generative adversarial networks (GANs) is analogous to two models being pitted against each other, where one model (the generator *G*) counterfeits for example images and the other model (the discriminator *D*) estimates the probability for whether they are fake or not. This competition ideally drives both models to improve until fake images become indistinguishable from real ones and is formulated as a minimax game [7]. GANs have proven enormously successful in generative applications and were extended to conditional settings to solve ill-posed problems such as text-to-image translation [8], image-to-image translation [1] or single image super-resolution [9]. Conditional GANs receive, alongside *z*, an additional non-random input to condition on.

For this reason they are closely related to the task of semantic segmentation with *D* being interpretable as a learned higher-order loss [1, 2]. The present de facto standard training loss for semantic segmentation is a multi-class cross-entropy loss $\mathcal{L}_{\text{mce}}$ that penalizes each pixel independently from all others:

$$\mathcal{L}_{\text{mce}}(\boldsymbol{\theta}_S) = -\frac{1}{N \cdot M} \sum_i^N \sum_j^M \boldsymbol{y}_{i,j}^{\mathsf{T}} \log S_j(\boldsymbol{x}_i), \quad (1)$$

where *j* runs across pixels in the instances *i* of a minibatch. In the adversarial training scheme, by contrast, the discriminator *D* classifies the segmentation quality on the basis of how well the whole label map fits the given image and is trained alongside the segmentor *S*. In the spirit of the minimax game, *D*'s loss can be formulated as follows:

$$\mathcal{L}_D(\boldsymbol{\theta}_D, \boldsymbol{\theta}_S) = -\frac{1}{N} \sum_i^N \Big( \log D\left(\boldsymbol{y}_i, \boldsymbol{x}_i\right) + \log \left(1 - D\left(S\left(\boldsymbol{x}_i\right), \boldsymbol{x}_i\right)\right) \Big) \quad (2)$$

with *S* in turn minimizing $\mathcal{L}_S = -\mathcal{L}_D$. We however follow [7] and use the loss-term below for the sake of larger gradient signals:

$$\mathcal{L}_S(\boldsymbol{\theta}_D, \boldsymbol{\theta}_S) = -\frac{1}{N} \sum_i^N \log D\left(S\left(\boldsymbol{x}_i\right), \boldsymbol{x}_i\right) \quad (3)$$

Luc *et al.* [2] propose a hybrid loss term for the segmentor *S* in form of a weighted sum, $\mathcal{L}'_S(\boldsymbol{\theta}_D, \boldsymbol{\theta}_S) = \mathcal{L}_S + \lambda \mathcal{L}_{mce}$, which we also compare against below. Optimal training requires for *D* to be near its optimal solution at all times. For this purpose, *D* can be trained using *k* minibatch gradient descent steps for each such step performed on *S* [7].

### 2.2 MRI dataset

The employed dataset contains 152 patients with MRI acquired using a Siemens Prisma 3.0 T machine at the National Center for Tumor Diseases (NCT) in Heidelberg, Germany. Prospective data collection was approved by a local and governmental ethics committee and informed consent was obtained from all patients prior to entering this study. All patients had a suspicious screening result and a core biopsy yielding pathological classification, i.e. Gleason Score (GS) [10]. Image analysis was based on a T2-weighted Image (T2w), an Apparent Diffusion Coefficient (ADC) map and a high b-value diffusion weighted image (b1500) at $b = 1500 \, \text{s} \, \text{mm}^{-2}$. The T2w images have an in-plane resolution of $0.25 \, \text{mm}$, the other two modalities were upsampled accordingly. The prostate's anatomical details as well as lesions were segmented independently on both the T2w and the ADC-map by an experienced radiologist. The annotations comprise four classes: tumor lesion, peripheral zone, transitional zone and other (i.e. background). Registration was performed using rigid translation maximizing the overlap between the PZ masks.

### 2.3 Training

To provide meaningful comparison, the training protocol is the same for all evaluated schemes. We use a set of 55 patients ($\mathcal{S}_{agg}$) comprising 188 2D-slices with biopsy-confirmed aggressive tumor lesions of GS $\geq$ 7 and 97 patients ($\mathcal{S}_{free}$) with 475 2D-slices that were diagnosed lesion free (slice size 3×416×416). The experiments are performed using four-fold cross-validation on $\mathcal{S}_{agg}$ with mutually exclusive subject allocation to the folds, while $\mathcal{S}_{free}$ is used during training only. In each cross-validation permutation, 2 folds are employed for training the model, one fold for model selection according to the tumor dice, and one held-out fold for validation. All segmentation models are trained for 225 epochs, with 80 randomly sampled batches each, using an initial learning rate (LR) of $10^{-5}$, that is halved every 75 epochs. During the adversarial training scheme we train the discriminator $D$ on 3 batches for each batch the segmentor is trained on while using fixed LR $= 10^{-5}$ for $D$. For parameter optimizations we use *Adam* [11]. The training data is augmented by in-plane rotations with angle $\phi \sim \mathcal{U}\left[-\pi/8, \pi/8\right]$, crops with a mask shifted by $(\Delta x, \Delta y) \sim (\mathcal{U}\left[-50, 50\right], \mathcal{U}\left[-50, 50\right])$ and random left-right mirroring. We use a batch-size of 5 with importance sampling, averaging to 3.5 samples from $\mathcal{S}_{agg}$ in each batch.

### 2.4 Network Architectures

We use an identical 'U-Net'-type architecture for the segmentor in each experiment [12]. We follow [1] and use InstanceNorm instead of BatchNorm, conjecturing that it avoids harmful stochasticity, introduced by small batch-sizes. Let `Ck` denote a Convolution-InstanceNorm-ReLU layer. Then the segmentor's encoder takes on the following form: `CL64-CL128-CL256-CL512-CL1024`, while the decoder can be represented as: `C512-C256-C128- 64-C4`. The architecture used for the discriminator in large parts mirrors that of the segmentor's encoder: `CL64-CL128-CL256-CL512-CL512-CL1024-GPD1`, where `GPD1` denotes a global average pooling layer followed by a dense layer with one output node. InstanceNorm is neither applied to the first nor the last layer in $S$ and $D$. Convolutional layers employ 3×3-filters, except for the last one in $S$'s decoder which uses 1×1-filters. $D$ takes 7×416×416 inputs, featuring three channels for the MRI modalities and four channels encoding the class labels.

## 3 Results

The adversarial approach scored significantly better for tumor segmentation both in the Dice coefficient (DSC) as well as the sensitivity (Tab. 1, $p < 0.001$ using Wilcoxon signed-rank test). The specificities between the approaches were equal. Fig. 3 illustrates examplary segmentations. Using a hybrid loss with the same weighting as [2] does not provide further improvements.

Table 1: Experimental results of the four-fold cross-validation for GS $\geq$ 7 Tumor.

| training scheme<br>loss | cross-entropy<br>$\mathcal{L}_{mce}$ | adversarial<br>$\mathcal{L}_S \& \mathcal{L}_D$ | hybrid<br>$\mathcal{L}_{mce}/2 + \mathcal{L}_S \& \mathcal{L}_D$ |
|---|---|---|---|
| tumor DSC | $0.35 \pm 0.29$ | **$0.41 \pm 0.28$** | $0.39 \pm 0.29$ |
| tumor sensitivity | $0.37 \pm 0.33$ | **$0.55 \pm 0.36$** | $0.49 \pm 0.35$ |
| tumor specificity | $0.98 \pm 0.14$ | $0.98 \pm 0.14$ | $0.98 \pm 0.14$ |

In order to evaluate how the training schemes compare on progressively smaller datasets, we successively take away positive training samples from the fold that both schemes coincided to perform best on. We train in the exact same manner as described above and evaluate on the same held-out fold from before. The results are depicted in Fig. 2.

## 4 Discussion

The adversarial training scheme can be viewed as a learned reparametrization of the groundtruth that is employed to pass on back-propagation training signals to the segmentation network. This reparametrization allows for the co-existence of multiple plausible segmentations, i.e. the subspaces of equally correct segmentations are recognized as such, which is in contrast to the de facto standard loss formulation, the cross-entropy loss. The latter formulates a wrongfully definitive groundtruth that is also agnostic to long-range dependences and thus structural relationships in the label maps. Our experiments show that this can be harmful, especially in the medical domain, where label ambiguities are arguably a common sight. Our proposed method on the other hand is both, more data-efficient and increases detection sensitivity and the dice-score of aggressive prostate cancer. This segmentation task is challenging due to the strong tissue heterogeneities of the prostate and the subtle tumor appearance, leading to large intra- and inter-rater variability in the groundtruth.

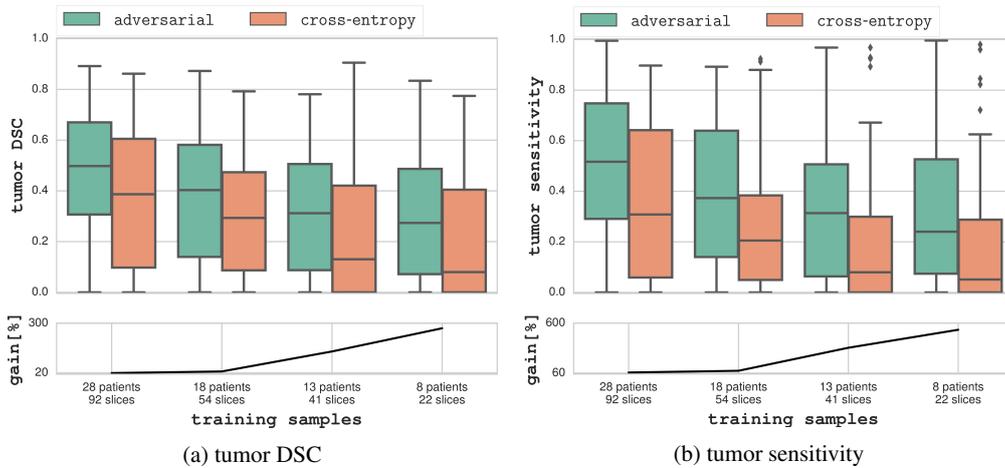

(a) tumor DSC  (b) tumor sensitivity

Figure 2: Comparison of performance in terms of DSC (a) and sensitivity (b) between the adversarial and cross-entropy training when successively taking away training data. The upper panels illustrate the respective distributions for the two schemes. The lower panels show the relative gain in median of the adversarial over the cross-entropy training, from which particularly pronounced gains are visible in the small dataset limit. Specificity (not shown) was around 0.98 in all experiments.

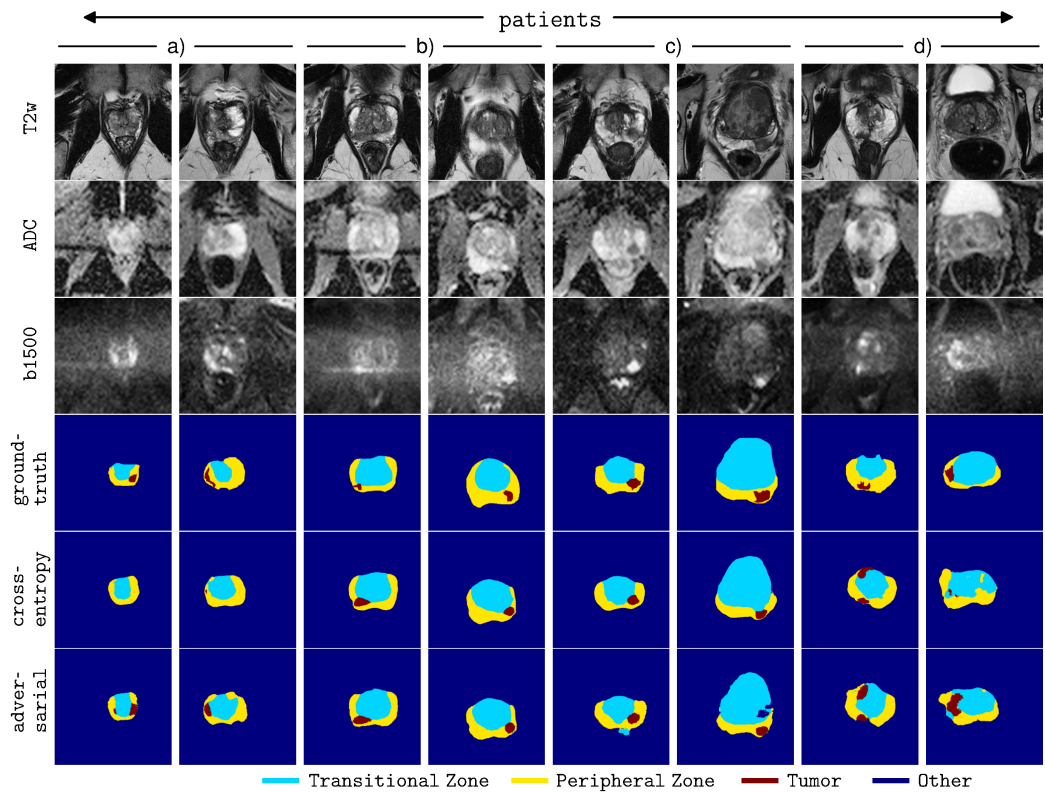

Figure 3: Examples depicting the three MRI modalities, the expert annotation as well as the segmentations produced by training the segmentor network with different loss schemes. The first two columns from the left, i.e. columns a), depict examples in which the adversarial is clearly more sensitive to aggressive tumor than the cross-entropy training. Columns b) show examples for which the methods are on par. Columns c) feature examples for which the adversarial method yields partially defective label maps. Columns d) exhibit examples for which both methods deviate considerably from the ground-truth, the first of which likely shows tumor detection by both methods, missed by the expert.

5